\documentclass[preprint,12pt]{elsarticle}

\usepackage{amssymb}
\usepackage{amsmath}
\graphicspath{ {./figures/} }
\usepackage{hyperref}
\usepackage{float}
\usepackage{verbatim} %comments
\usepackage{apalike}
\restylefloat{figure}
\floatstyle{plaintop} %table caption at top
\restylefloat{table}
\usepackage{booktabs}
\usepackage[table]{xcolor}
\journal{Knowledge-Based Systems}

\begin{document}

\begin{frontmatter}

%% Title, authors and addresses

%% use the tnoteref command within \title for footnotes;
%% use the tnotetext command for theassociated footnote;
%% use the fnref command within \author or \affiliation for footnotes;
%% use the fntext command for theassociated footnote;
%% use the corref command within \author for corresponding author footnotes;
%% use the cortext command for theassociated footnote;
%% use the ead command for the email address,
%% and the form \ead[url] for the home page:
%% \title{Title\tnoteref{label1}}
%% \tnotetext[label1]{}
%% \author{Name\corref{cor1}\fnref{label2}}
%% \ead{email address}
%% \ead[url]{home page}
%% \fntext[label2]{}
%% \cortext[cor1]{}
%% \affiliation{organization={},
%%             addressline={},
%%             city={},
%%             postcode={},
%%             state={},
%%             country={}}
%% \fntext[label3]{}

\title{Focus on What Matters: Constraining Spatial-Temporal Attention via Action-Units for Noise-Resilient AQA} %% Article title

%% use optional labels to link authors explicitly to addresses:
%% \author[label1,label2]{}
%% \affiliation[label1]{organization={},
%%             addressline={},
%%             city={},
%%             postcode={},
%%             state={},
%%             country={}}
%%
%% \affiliation[label2]{organization={},
%%             addressline={},
%%             city={},
%%             postcode={},
%%             state={},
%%             country={}}

\author[1]{Shuikang Zhu}
\author[2]{Yiding Sun}
\author[2]{Zihao Guo}
\author[1]{Yang Yang}
\author[3]{Chen Sun}

\affiliation[1]{organization={Faculty of Electronic and Information Engineering, Xi'an Jiaotong University}}

\affiliation[2]{organization={School of Software Engineering, Xi'an Jiaotong University}}

\affiliation[3]{organization={School of Intelligent Sports Engineering, Shanghai Sport University}}
\begin{abstract}
%% Text of abstract
The core challenge in Action Quality Assessment (AQA) lies in extracting fine-grained motion features from redundant and complex video backgrounds. Existing global feature learning methods are constrained by extremely low "signal-to-noise ratios", making it difficult to distinguish intrinsic actions from background clutter. To address this, we propose a Pose-Guided Intrinsic Motion Distillation Framework that explicitly enforces physical constraints to focus on motion subjects and decouple motion execution from environmental outcomes. First, we design an Action-Unit Parser that constructs dynamic regions of interest (ROIs) using human pose topology as prior knowledge. This functions as a spatial hard-attention filter that physically removes background noise at the input stage, forcing the model to learn appearance and geometric features only from pure body regions. Second, to resolve factor entanglement, we introduce a dual-stream decoupling mechanism: the Motion Parser focuses on capturing purified joint motion details, while the Condition Parser independently processes non-body-related environmental feedback (e.g., splash in diving) to create two orthogonal evaluation dimensions in feature space. Finally, adaptive weight modules integrate these decoupled features to generate final scores. Experimental results on large-scale datasets including FineDiving, FineDiving-HM, and MTL-AQA demonstrate that this method achieves state-of-the-art (SOTA) performance in both action segmentation and scoring accuracy, validating the effectiveness of "noise suppression focusing" and "motion disentanglement" strategies in fine-grained action evaluation.

\end{abstract}

%%Graphical abstract
\begin{graphicalabstract}
\includegraphics[width=1\linewidth]{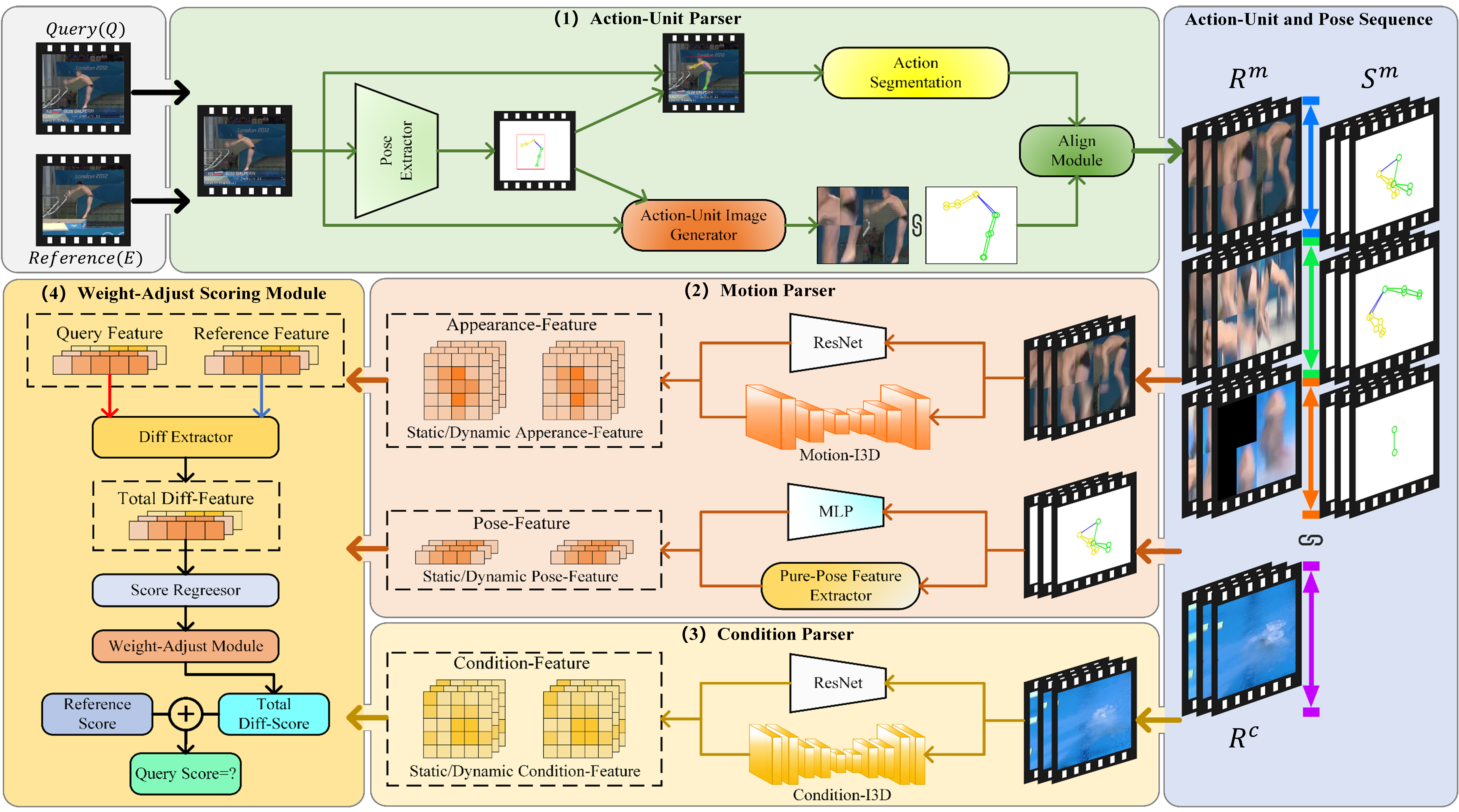}

\end{graphicalabstract}

%%Research highlights
\begin{highlights}
\item A pose-guided hard-attention filter physically eliminates background noise.
\item Action-Units enforce spatial constraints to focus on intrinsic motion regions.
\item Dual-stream decoupling separates motion execution from environment feedback.
\item Geometric pose priors improve both scoring accuracy and action segmentation.
\item State-of-the-art results achieved on FineDiving and MTL-AQA datasets.
\end{highlights}

%% Keywords
\begin{keyword}
%% keywords here, in the form: keyword \sep keyword
Action Quality Assessment \sep Pose-Guided Hard-Attention \sep Motion Decoupling 

%% PACS codes here, in the form: \PACS code \sep code

%% MSC codes here, in the form: \MSC code \sep code
%% or \MSC[2008] code \sep code (2000 is the default)

\end{keyword}

\end{frontmatter}
\section{Introduction}
\label{introduction}

\begin{figure}[t]
    \centering
    \includegraphics[width=1\linewidth]{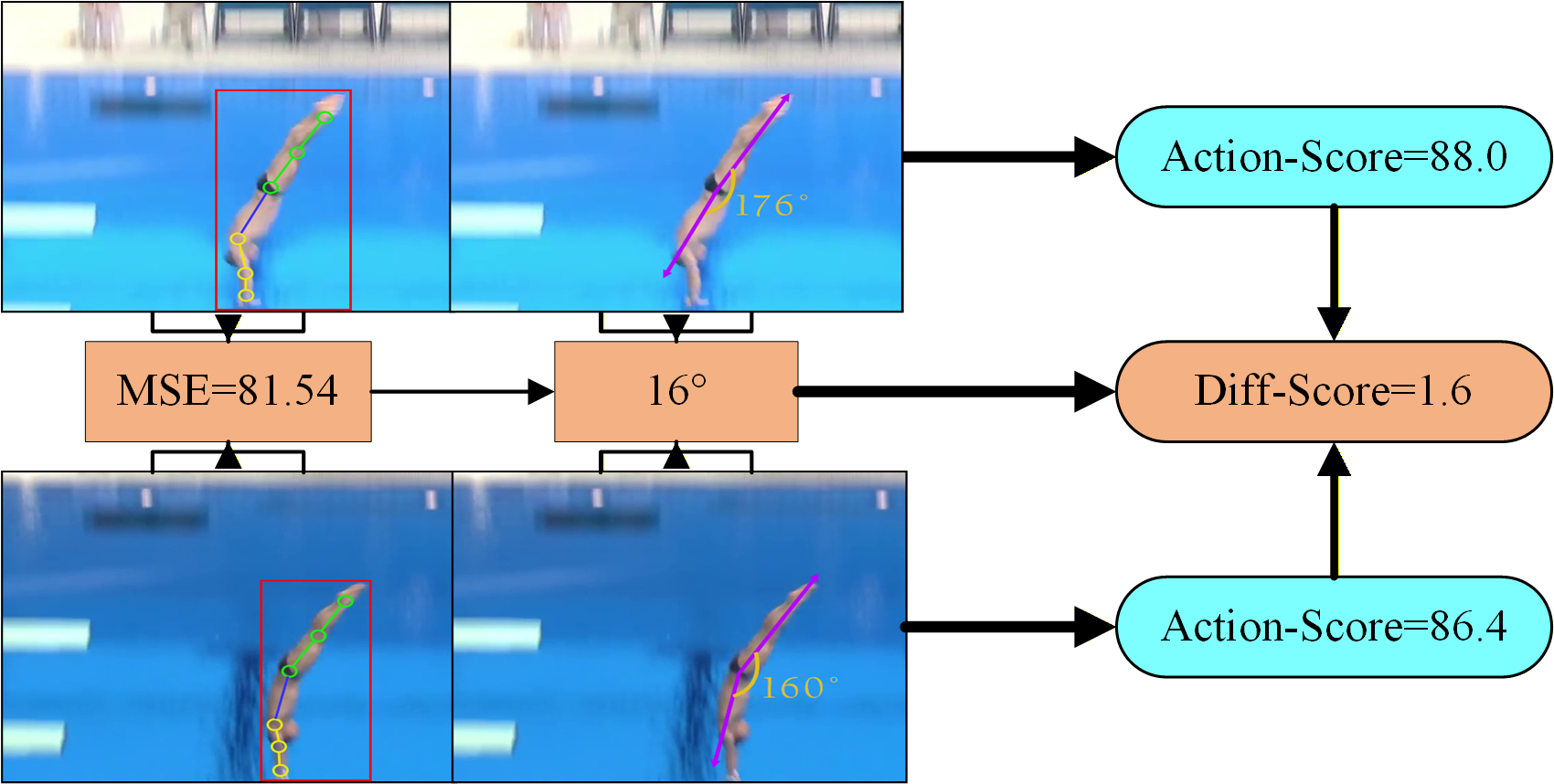}
    \caption{
        An illustration of the sensitivity of Action Quality Assessment (AQA) to subtle kinematic variations. As shown, a minor structural difference of $16^\circ$ in the body bending angle between two diving executions results in a noticeable score penalty of 1.6 points. This highlights the critical necessity for models to accurately capture fine-grained pose features rather than relying solely on global appearance.
    }
    \label{fig:motivation}
\end{figure}

\noindent Action Quality Assessment (AQA) aims to quantify the refinement of action execution, holding significant value in sports scoring, rehabilitation training, and medical evaluation \cite{art-46,art-47,art-48}. Unlike action recognition tasks that focus on identifying "what action occurs" \cite{art-51,art-52,art-53,art-55}, AQA's core challenge lies in discerning "how well the action was performed". This requires models to capture subtle kinematic features (e.g., knee bending angles, entry postures), as explicitly illustrated in Figure \ref{fig:motivation}. However, existing AQA methods \cite{art-16,art-6,art-27} predominantly adopt full-frame video inputs, which introduce a significant "low signal-to-noise ratio" (SNR) problem in complex sports scenarios: background clutter (e.g., spectators, billboards) often occupies most of the visual field, while essential limb movements determining scores constitute only a minor portion. This information asymmetry leads deep models to easily fall into "shortcut learning" - rating based on environmental context rather than motion quality - severely compromising evaluation objectivity and robustness. Formally, this can be characterized as a distributional shift problem, where the model's prediction spuriously correlates with the environmental covariate rather than the causal motion feature. This spurious correlation fundamentally contradicts the core philosophy of AQA, which mandates that the score should be a deterministic function of the performer's kinematics alone.\\
\indent To break through this bottleneck, we argue that it is essential to restructure the model's acquisition of visual information at the physical level. This paper proposes a Pose-Guided Intrinsic Motion Distillation Framework (PGIMDF) that aims to enforce the model to regress to the intrinsic motion through "noise reduction focusing" and "orthogonal decoupling" strategies \cite{art-56}.\\
\indent Our core insight is that human pose is not merely a set of features but a natural "spatial hard-attention filter". Unlike previous implicit soft-attention mechanisms, we construct dynamic Action Units using pose topology to physically eliminate background noise at the input stage. This creates a spatiotemporal channel containing only human regions, eliminating the reliance on background information while focusing on learning pure limb appearance and geometric evolution. Furthermore, to address the complexity of scoring factors, we design a three-decoupled architecture (Pose-ROI-Condition). By employing independent parsers to separately process limb movements (e.g., somersaults) and non-limb environmental feedback (e.g., splash effects), we establish two orthogonal evaluation dimensions in feature space, achieving independent and fair measurement of action execution and environmental outcomes.\\
\indent This paper's main contributions are as follows: \\
\indent (1) We propose a noise-robust AQA framework based on hard attention, which enables the model to focus more on subtle pose variations related to motion, significantly enhancing its understanding of complex actions.\\
\indent (2) We design a dual-stream decoupling mechanism for motion and environment, separating action execution from environmental feedback from a physically interpretable perspective, aligning closer with human judges' cognitive processes.\\
\indent (3) Experimental results on large-scale datasets including FineDiving \cite{art-16}, FineDiving-HM \cite{art-12}, and MTL-AQA \cite{art-15} demonstrate that our method achieves state-of-the-art (SOTA) performance in both action segmentation and scoring accuracy.
\begin{figure}[t]
	\centering
	\includegraphics[width=1\linewidth]{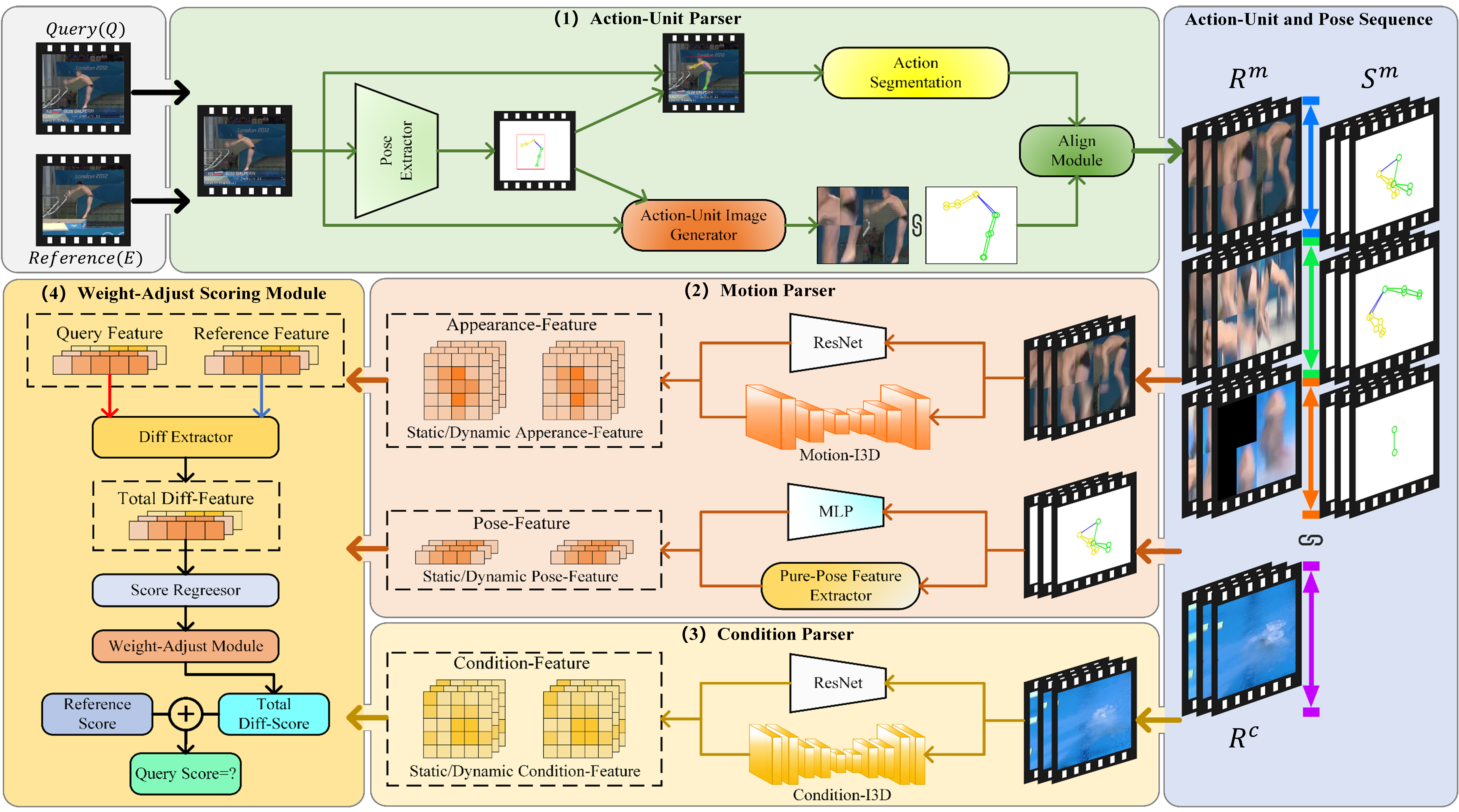}
	\caption{
		The structure of a Pose-Guided Intrinsic Motion Distillation Framework is proposed. We utilize a multi-level parser to separate foreground athletes from the input video pair, extract pose information, and generate Action Units. Then, we obtain appearance features, pose features, and condition features by these parser. By comparing the differences in these features, we ultimately regress to the score difference.
	}
	\label{fig:framework}
\end{figure}
\section{Related Work}
\subsection{Action Quality Assessment (AQA)}
\noindent The AQA task aims to eliminate subjective biases in sports scoring \cite{art-5}. Early work relies on handcrafted features or shallow regression models \cite{art-1,art-2}, struggling to capture complex spatiotemporal dynamics. With the advancement of deep learning, direct regression methods based on C3D \cite{art-3}, I3D \cite{art-13}, and LSTM \cite{art-4} significantly enhance motion trajectory modeling \cite{art-15,art-22,art-29}. To address subtle differences, Yu et al. \cite{art-6} introduce pairwise contrastive learning, while An et al. \cite{art-30} propose a multi-stage comparison regression framework. However, these methods primarily adopt full-frame (Full-frame) inputs, suffering from low signal-to-noise ratio issues. Background elements such as spectators and billboards often dominate feature extraction, overshadowing motion details. FineParser \cite{art-12} attempts to remove background pixels using foreground masks, but this remains essentially a pixel-level "soft filtering". Recently, Qi et al. \cite{art-54} try replacing mask features with pose features for feature fusion on FineParser. However, the performance degradation compared to FineParser indicates that the model has not yet captured the deep features inherent in motion-based pose variations. Compared to existing methods, the proposed framework constructs physical-level hard attention through pose topology, not only effectively suppressing background noise but also achieving explicit modeling of motion geometry structures.
\subsection{Action Segmentation}
\noindent Action segmentation aims to provide temporal boundaries for fine-grained analysis \cite{art-9,art-10,art-11}. Traditional methods rely on LSTM \cite{art-4} or Transformer \cite{art-7} to capture inter-frame correlations. To address over-segmentation, ASRF \cite{art-33} introduces boundary detection mechanisms, while MS-TCN \cite{art-32} employs multi-level temporal convolutions to expand receptive fields. Some studies attempt to enhance feature representation by incorporating region attention \cite{art-8} or graph convolutional networks \cite{art-31}. However, these methods often underperform in sports scenarios. The reason lies in their reliance on global appearance features for phase transition detection, where visual features become ambiguous in fast motion or complex backgrounds. In contrast, we introduce human pose into segmentation tasks, utilizing geometric changes (rather than texture changes) in skeleton sequences as physical anchors for action boundary determination. This approach achieves more robust segmentation performance in high-dynamic scenarios.
\section{Method}
\noindent The structure of our framework is shown in Figure \ref{fig:framework}.
It is driven by four core components: (1) The Action-Unit Parser localizes foreground athletes and utilizes pose topology to generate Action Units frame-by-frame; (2, 3) The Motion Parser and Condition Parser form a dual-stream decoupling architecture, which collaboratively distills appearance, pose, and condition features from the generated Action Units; (4) The Weight-Adjust Scoring Module integrates these decoupled features to regress the final score. The following subsections detail the design of each component.
\subsection{Problem Formulation}
\noindent Given a pair of query video $Q$ and reference video $E$ with the same action type, our framework aims to predict the score of Q. Since contrastive methods \cite{art-6,art-12} demonstrate significant effectiveness, we adopt this method to compare subtle differences in pose information and obtain score differences that can be integrated into the final score. The overall framework is formulated as follows:
\begin{equation}
	{\hat{X}}_Q=\mathcal{F}\left(Q,E\right)+X_E,
	\label{eq:1}
\end{equation}
where ${\hat{X}}_Q$ denotes the predicted score of query video $Q$, and $X_E$ represents the true score of reference video $E$.
\begin{figure}[t]
	\centering
	\includegraphics[width=1\linewidth]{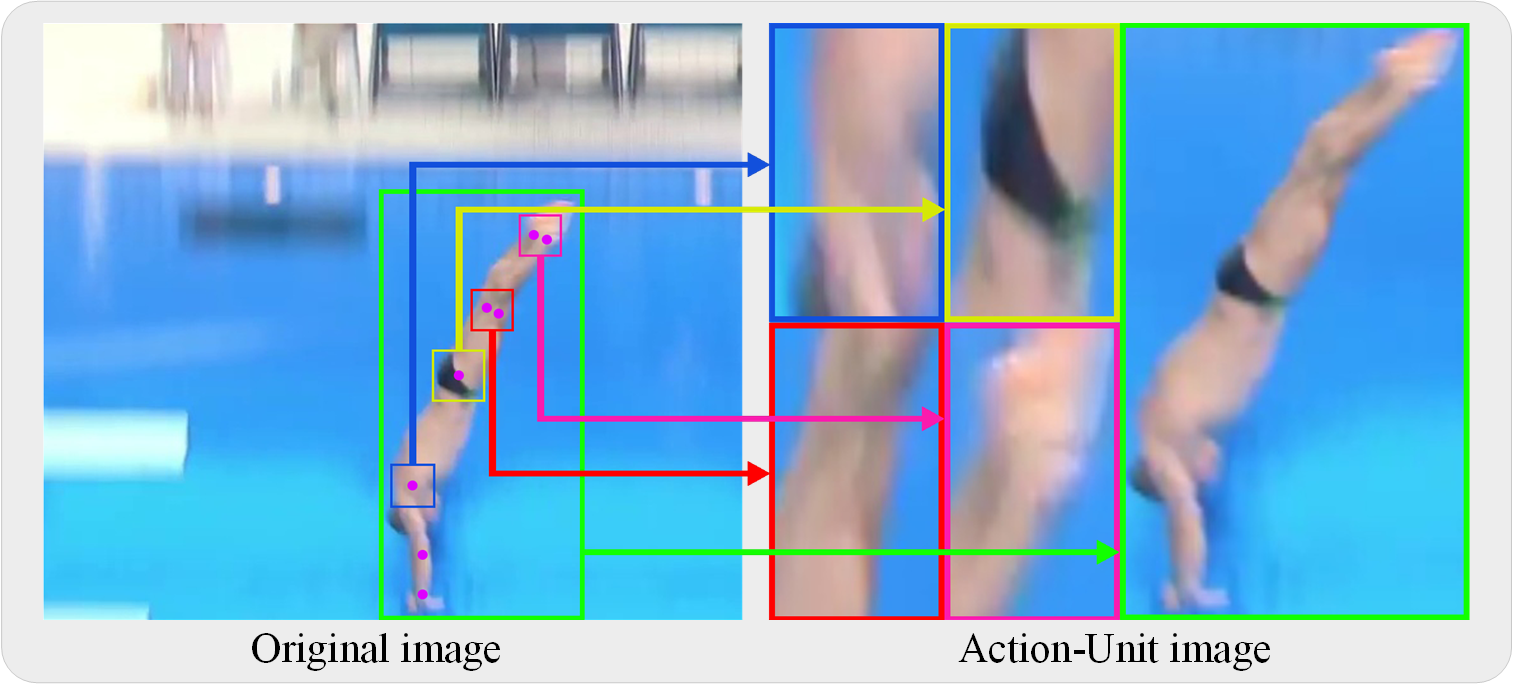}
	\caption{
		The visualization of the composition of the Action-Unit image.
	}
	\label{fig:action-unit}
\end{figure}
\subsection{Action-Unit Parser}
\noindent This is the core component that enforces physical-level constraints on data, performing three main tasks: (1) extract and utilize full pose information (including 2D bounding boxes, 2D joint coordinates, body bending angles); (2) generate Action-Unit images; and (3) conduct action segmentation and alignment. The following content details how each sub-module collaborates to complete these tasks.
\paragraph{\textbf{Pose Extractor}} The extractor primarily comprises a top-down 2D pose estimation module. First, it extracts frame-by-frame 2D bounding boxes and 2D joint coordinates corresponding to foreground athletes in the input video. Then, it calculates bending angles for several predefined body regions based on these 2D joint coordinates. Specifically, the selected body regions may vary depending on the type of motion. In this paper, we select ankle, knee, waist, and shoulder joint bending angles for calculation. Finally, all 2D joint coordinates and corresponding body bending angles are integrated into a pose information sequence for output, denoted as $S$. Similarly, the bounding box sequence is also outputted, denoted as $B$.
\begin{figure}[t]
	\centering
	\includegraphics[width=1\linewidth]{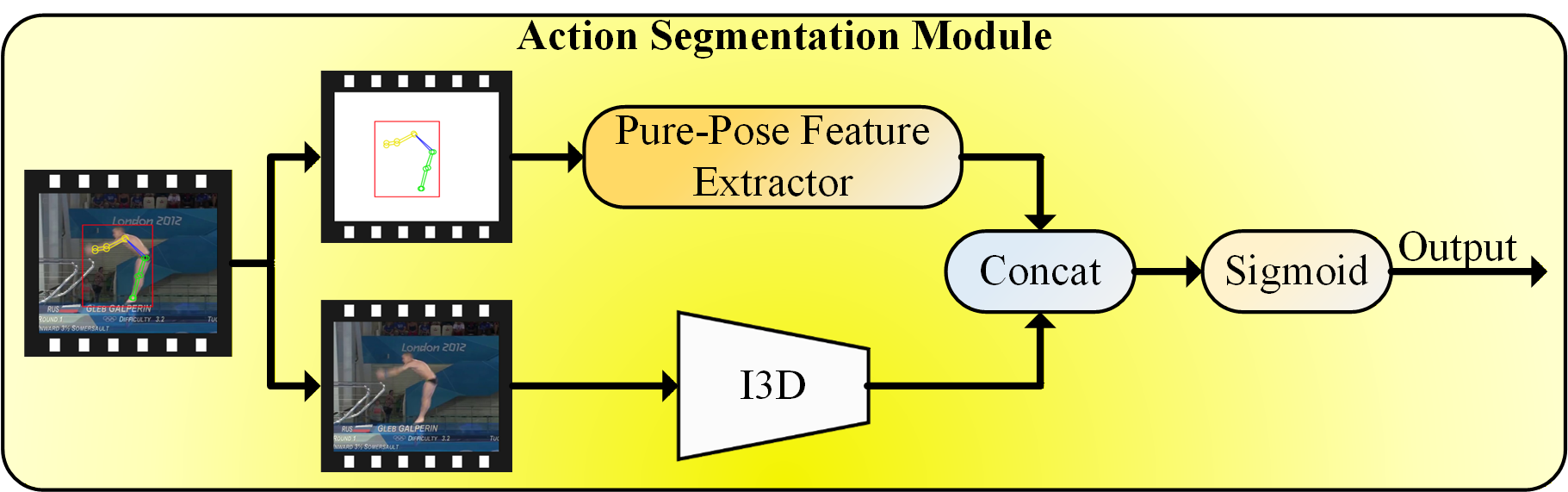}
	\caption{
		The network structure of the Action Segmentation Module.
	}
	\label{fig:action-segmentation}
\end{figure}
\paragraph{\textbf{Action Segmentation Module}} This sub-module is used to divide a complete motion into several sub-phases for detailed feature comparison across each phase. Its structure is shown in Figure \ref{fig:action-segmentation}. First, we assume that we need to identify the frame index positions for $H$ action transitions. The I3D network captures the action transition features embedded in the video frame sequence with the help of pose information, and predicts the probability of action transition occurring in the $t_{th}$ frame of the sports video based on this feature. The specific representation is as follows:
\begin{equation}
	{\hat{P}}_Q=\left\{\left[{\hat{p}}_1^h,{\hat{p}}_2^h,\ldots,{\hat{p}}_T^h\right]\right\}_{h=1}^H=ASM\left(Q,S_Q\right),
	\label{eq:2}
\end{equation}
\begin{equation}
	{\hat{K}}_Q^h=argmax({\hat{P}}_Q),
	\label{eq:3}
\end{equation}
where $ASM$ refers to Action Segmentation Module. ${\hat{P}}_Q$ is a probability matrix with shape $T\times H$, where the $t_{th}$ value in the $h_{th}$ row represents the probability of the $h_{th}$ action transition occurring in the $t_{th}$ frame, and $k_h$ denotes the frame index of the $h_{th}$ action transition. $K_Q=\{{k_Q^i}\}_{i=1}^H$ represents the set of keyframes for the query video.
\paragraph{\textbf{Action-Unit Image Generator}} The generator generates action interest region images corresponding to each frame based on $Y$ joint points specified in advance and the corresponding $S$ and $B$. The combination method of Action-Unit images is shown in Figure \ref{fig:action-unit}. These images are referred to as Action-Unit images in this paper and denoted by $o$. The set of Action-Unit images corresponding to the entire video frame set is denoted as $R=\{o_t\}_{t=1}^T$.
\paragraph{\textbf{Align Module}} This module executes three steps: (1) it separates the pose information sequence ($S$) and Action-Unit image set ($R$) into the motion part (the complete action of the athlete) and the condition part (special score influencing factors such as water splash and height above ground, etc.) through the keyframe set ($K$); (2) it further divides the motion part into $H+1$ sub-phases; (3) to ensure a fair comparison between the query video $Q$ and the reference video $E$, we perform frame alignment for each sub-phase of their motion parts and overall alignment for the condition parts. Specifically, the data for the condition part is represented as $R^c$, while the data for the motion part is represented as follows:
\begin{equation}
	S^m,R^m=\bigcup_{i=1}^{H+1}{S^{m,i},R^{m,i}},
	\label{eq:4}
\end{equation}
where $S^{m,i}$ represents the pose information of the $i_{th}$ sub-phase of the motion part, while $R^{m,i}$ represents the set of Action-Unit images corresponding to the sub-phase. The aligned data are denoted as: ${\bar{S}}^{m,i}$, ${\bar{R}}^{m,i}$, and ${\bar{R}}^c$.

\begin{figure}[t]
	\centering
	\includegraphics[width=1\linewidth]{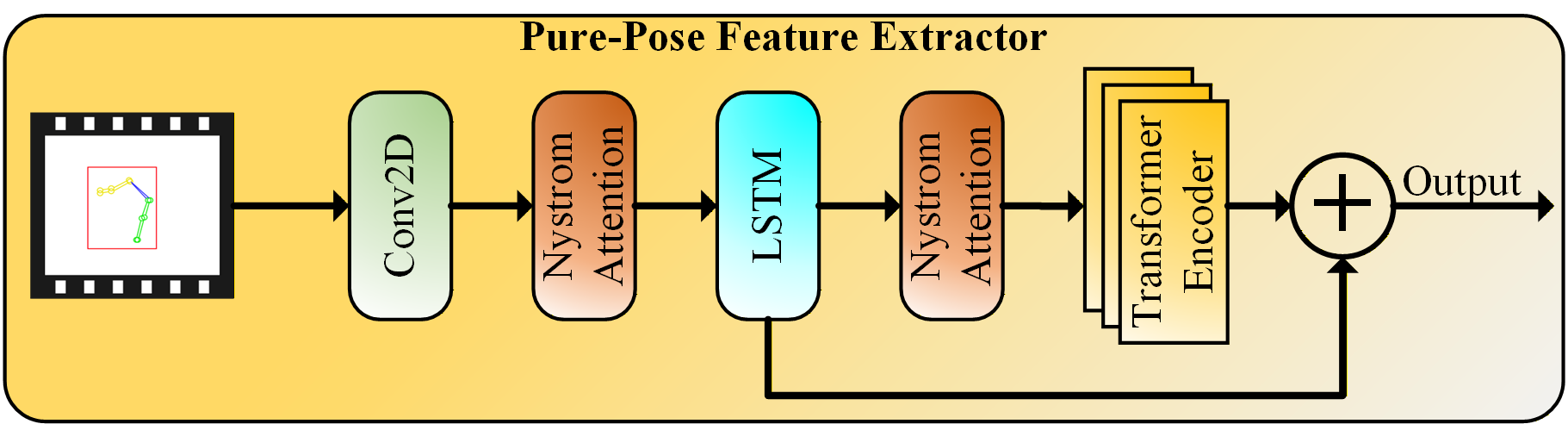}
	\caption{
		The network structure of the Pure-Pose Feature Extractor.
	}
	\label{fig:pure-pose}
\end{figure}
\subsection{Feature Parser}
\noindent Within this framework, we design a dual-stream decoupling architecture consisting of the Motion Parser and the Condition Parser. While structurally divided into two streams, they collaboratively distill three distinct types of features—appearance, pose, and condition—to achieve a comprehensive Pose-ROI-Condition decoupled evaluation. Specifically, the dynamic branches of these parsers are responsible for extracting dynamic appearance features ($F^a$), dynamic pose features ($F^p$), and dynamic conditional features ($F^c$). Building upon the good practices from \cite{art-12}, we also set branches to extract their static versions for each feature. These static branches output three types of static features (${SF}^a$, ${SF}^p$ and ${SF}^c$). In the following, we introduce the method by which the Motion Parser extracts pose features.\\
\indent In the Motion Parser, pose feature extraction is performed by the Pure-Pose Feature Extractor. Its architecture is shown in Figure \ref{fig:pure-pose}. First, it is important to note that the Conv2D layer establishes associations between coordinates of different joints and body joint bending angles through learnable parameters. This enables the model to autonomously construct the most suitable pose topology for the current scenario. The design is partially motivated by recent self-supervised representation learning paradigms \cite{art-57}, which emphasize the importance of capturing intrinsic geometric structures from raw data.\\
\indent Furthermore, in terms of temporal feature processing, this structure that concatenates a Transformer encoder after an LSTM effectively addresses the performance limitations of individual models by integrating time-dynamic modeling and global dependency capture capabilities: (1) Although traditional LSTM captures local temporal dependencies through gating mechanisms, it is limited by the insufficient modeling ability of the recurrent structure for long-distance dependencies; (2) Although Transformer implicitly introduces temporal information through positional encoding to model global dependencies, it has an efficiency bottleneck in the extraction of local dynamic features in strong temporal data. Additionally, we further enhance the module's ability to extract contextual features by leveraging the Nystrom Attention \cite{art-38} while maintaining a minimal parameter count.
\subsection{Weight-Adjust Scoring Module}
\noindent To accommodate the varying requirements of different sports events regarding the relative importance of motion and condition parts, we propose a scoring module with customizable weights. For example, in international diving competitions, the execution (splash) accounts for 30-40\% of the total score, while body movements account for 60-70\%. Furthermore, judges may assign different weightings to different sub-stages of the motion part (e.g., take-off accounting for 30\%, turning for 55\%, and entry for 15\%). Our module allows the contribution of each component to be flexibly adjusted according to the specific scoring rules of each sport.\\
\indent The module first utilizes the Diff Extractor to extract the differential information between features—namely $F^a$, $F^c$, and $F^p$, along with their corresponding static versions—obtained from two videos.\\
\indent Secondly, the Score Regressor is employed to estimate the score differences associated with the differential information from various features. In official competitions, the weightings for motion part and condition part are generally fixed. However, the weightings for individual sub-stages within the motion part are often influenced by subjective factors of judges. To control the weighted proportions between motion part score and condition part score, we set two prior hyperparameters ($\alpha$ and $\beta$). For the weightings of sub-stages in the motion part, we assign learnable parameters to each stage and apply softmax at the end to ensure they conform to a probability distribution. The total score difference is then obtained by aggregating these weighted differences across all stages of the movement. Given the presence of two distinct feature types—dynamic and static—we further fuse the respective score differences in a 5:5 ratio to produce the final score differences. The score difference is the final output of the entire framework, and the predicted score of the query video is obtained by adding the real scores of the reference video.
\subsection{Training and Inference}
\paragraph{\textbf{Loss-function}} We select the query video $Q$ and the reference video $E$ from the training set. We optimize the entire framework by minimizing the following loss function.
\begin{equation}
	\mathcal{L}=\mathcal{L}_{ASM}+\mathcal{L}_{MSE}.
	\label{eq:5}
\end{equation}
\indent $\mathcal{L}_{ASM}$ is used to optimize the Action Segmentation Module, and its calculation method is as follows:
\begin{equation}
	\mathcal{L}_{ASM}=\sum_{h=1}^{H}\sum_{t=1}^{T}BCE\left(p_t^h,{\hat{p}}_t^h\right).
	\label{eq:6}
\end{equation}
\indent Let $k_h$ denote the true frame index for the $h_{th}$ action transition, and $p^h$ be represented as a binary distribution, where ${p_t^h|}_{t\neq k_h}=0$ and ${p_t^h|}_{t=k_h}=1$. Conversely, ${\hat{p}}_t^h$ represents the predicted probability of the $h_{th}$ action transition occurring in the $t_{th}$ frame. $\mathcal{L}_{MSE}$ is used to optimize the entire action quality assessment model, and its calculation method is as follows:
\begin{equation}
	\mathcal{L}_{MSE}=|X_Q-{\hat{X}}_Q|^2.
	\label{eq:7}
\end{equation}
\paragraph{\textbf{Reference Video Selection Method}} We maximize the generalization performance of the model by introducing a multi-process variable $ReferenceDict$ during model training. This variable is used to store a list of video names that have already undergone comparative training with each query video. During each training epoch, videos that are not present in the list (i.e., videos that have not yet undergone comparative training) are selected for comparative training.
\paragraph{\textbf{Inference}} For the test video $Q_{test}$, we employ a multi-instance balanced voting mechanism \cite{art-6} to select $L$ reference videos from the training set for inference voting (i.e., $E_{test}=\{{E_l}\}_{l=1}^L$). These $L$ videos are then fed into the model for score inference, and the average of the $L$ scores is taken as the final prediction score for $Q_{T}$. The entire process can be expressed as follows:
\begin{equation}
	{\hat{X}}_{Q_{test}}=\frac{\sum_{l=1}^{L}\left(\mathcal{F}\left(Q_{test},E_l\right)+X_{E_l}\right)}{L}.
	\label{eq:8}
\end{equation}
\begin{table}[!t]
	\centering
	\caption{Performance comparison with state-of-the-art AQA methods on the FineDiving dataset. Our results are highlighted in bold format.}
	\begin{tabular}{l|cc}
		\toprule
		\multicolumn{3}{c}{AQA Task} \\
		\midrule
		Methods & $\rho\uparrow$ & $R_{\ell2}\downarrow(\times100)$ \\
		\midrule
		CoRe \cite{art-6} & 0.8631 & 0.3615 \\
		TSA \cite{art-16} & 0.9203 & 0.3420 \\
		GDLT \cite{art-24} & 0.9351 & 0.2684 \\
		$T^2$CR \cite{art-26} & 0.9382 & 0.2497 \\
		HGCN \cite{art-22} & 0.9381 & 0.2421 \\
		DAE \cite{art-25} & 0.9356 & 0.2493 \\
		CoFInAl \cite{art-27} & 0.9317 & 0.2887 \\
		HP-MCoRe \cite{art-54} & 0.9365 & 0.2440 \\
		\rowcolor[rgb]{.91, .91, .91} \textbf{Ours} & \textbf{0.9465} & \textbf{0.2243} \\
		\midrule
		\multicolumn{3}{c}{Action Segmentation Task} \\
		\midrule
		Methods & \multicolumn{2}{c}{AIoU@0.5/0.75$\uparrow$} \\
		\midrule
		ASFormer \cite{art-23} & 0.9913 & 0.8971 \\
		TSA \cite{art-16} & 0.8913 & 0.3822 \\
		NS-AQA \cite{art-17} & 0.8978 & 0.5733 \\
		HP-MCoRe \cite{art-54} & 0.9946 & 0.9718 \\
		\rowcolor[rgb]{.91, .91, .91} \textbf{Ours} & \textbf{0.9998} & \textbf{0.9841} \\
		\bottomrule
	\end{tabular}
	\label{tab:ex-finediving}
\end{table}
\section{Experiments}
\subsection{Dataset}
\paragraph{\textbf{FineDiving}} It comprises 3000 diving videos, encompassing 52 action types, 29 sub-action types, and 23 difficulty levels \cite{art-16}. It provides precise temporal boundaries and official action scores. Each action is identified by a specific number (DN), such as "407C" representing an inward dive of a half-flip. The dataset is divided into 2250 training samples and 750 testing samples, laying a solid foundation for accurate action recognition and analysis.

\begin{table}[!t]
	\centering
	\caption{Performance comparison with the state-of-the-art AQA methods on the FineDiving-HM dataset. Our results are highlighted in bold format. All methods utilize the unique mask annotation information of the FineDiving-HM dataset.}
	\begin{tabular}{l|cc}
		\toprule
		\multicolumn{3}{c}{AQA Task} \\
		\midrule
		Methods & $\rho\uparrow$ & $R_{\ell2}\downarrow(\times100)$ \\
		\midrule
		CoRe \cite{art-6} & 0.9308 & 0.3148 \\
		TSA \cite{art-16} & 0.9324 & 0.3022 \\
		FineParser \cite{art-12} & 0.9424 & 0.2602 \\
		\rowcolor[rgb]{.91, .91, .91} \textbf{Ours} & \textbf{0.9503} & \textbf{0.2179} \\
		\midrule
		\multicolumn{3}{c}{Action Segmentation Task} \\
		\midrule
		Methods & \multicolumn{2}{c}{AIoU@0.5/0.75$\uparrow$} \\
		\midrule
		TSA \cite{art-16} & 0.9239 & 0.5007 \\
		FineParser \cite{art-12} & 0.9946 & 0.9467 \\
		\rowcolor[rgb]{.91, .91, .91} \textbf{Ours} & \textbf{0.9999} & \textbf{0.9901} \\
		\bottomrule
	\end{tabular}
	\label{tab:ex-finediving-hm}
\end{table}
\paragraph{\textbf{FineDiving-HM}} It is an extended version of FineDiving, containing 312,256 masked frames covering all 3,000 videos \cite{art-12}.
\begin{table}[t]
	\centering
	\caption{Performance comparison with state-of-the-art AQA methods on the MTL-AQA dataset. Our results are highlighted in bold format.}
	\begin{tabular}{l|cc}
		\toprule
		\multicolumn{3}{c}{AQA Task} \\
		\midrule
		Methods & $\rho\uparrow$ & $R_{\ell2}\downarrow(\times100)$ \\
		\midrule
		CoRe \cite{art-6} & 0.9512 & 0.2600 \\
		TSA \cite{art-16} & 0.9422 & / \\
		GDLT \cite{art-24} & 0.9395 & 0.3990 \\
		$T^2$CR \cite{art-26} & 0.9529 & 0.2735 \\
		HGCN \cite{art-22} & 0.9522 & 0.2815 \\
		DAE \cite{art-25} & 0.9497 & 0.2869 \\
		CoFInAl \cite{art-27} & 0.9461 & 0.3461 \\
		\rowcolor[rgb]{.91, .91, .91} \textbf{Ours} & \textbf{0.9612} & \textbf{0.2452} \\
		\bottomrule
	\end{tabular}
	\label{tab:ex-mtl}
\end{table}
\paragraph{\textbf{MTL-AQA}} The multi-task action quality assessment dataset \cite{art-15} is collected from 1412 samples from 16 international top-level competitions. The annotation system covers multidimensional information such as difficulty level, individual scores from 7 judges, diving action categories, and final scores.
\begin{table}[t]
	\centering
	\caption{Generalization performance across different datasets.}
	\begin{tabular}{c|c|cc}
		\toprule
		Training dataset & Testing dataset & $\rho\uparrow$ & $R_{\ell2}\downarrow(\times100)$ \\
		\midrule
		MTL-AQA \cite{art-15} & FineDiving \cite{art-16} & 0.9005 & 0.3601 \\
		\bottomrule
	\end{tabular}
	\label{tab:cross-experiment}
\end{table}
\subsection{Evaluation Metrics}
\begin{table}[t]
	\centering
	\caption{Ablation study on different modules in our framework on the FineDiving dataset. The remaining components are adopted by default. $PE$ denotes Pose Extractor, $AIG$ denotes Action-Unit Image Generator, $PFP$ denotes Pure-Pose Feature Extractor, $AS$ denotes Action Segmentation Module.}
	\begin{tabular}{c|cccc|cc}
		\toprule
		Methods & $PE$ & $AIG$ & $PFP$ & $AS$ & $\rho\uparrow$ & $R_{\ell_2}\downarrow(\times100)$ \\
		\midrule
		A & $-$ & $-$ & $-$ & $-$ & 0.9217 & 0.3402 \\
		B & $-$ & $-$ & $-$ & $\boldsymbol{\checkmark}$ & 0.9251 & 0.3214 \\
		C & $\boldsymbol{\checkmark}$ & $-$ & $-$ & $\boldsymbol{\checkmark}$ & 0.9301 & 0.3001 \\
		D & $\boldsymbol{\checkmark}$ & $\boldsymbol{\checkmark}$ & $-$ & $\boldsymbol{\checkmark}$ & 0.9378 & 0.2665 \\
		E & $\boldsymbol{\checkmark}$ & $-$ & $\boldsymbol{\checkmark}$ & $\boldsymbol{\checkmark}$ & 0.9324 & 0.2897 \\
		\rowcolor[rgb]{0.91, 0.91, 0.91} \textbf{F} & $\boldsymbol{\checkmark}$ & $\boldsymbol{\checkmark}$ & $\boldsymbol{\checkmark}$ & $\boldsymbol{\checkmark}$ & \textbf{0.9465} & \textbf{0.2243} \\
		\bottomrule
	\end{tabular}
	\label{tab:main-ablation}
\end{table}
\noindent In this experiment, we inherit and extend the research findings of predecessors \cite{art-14,art-15,art-6,art-12}, selecting three commonly used and effective evaluation metrics to assess the accuracy and performance of the model in fine-grained action parsing tasks, namely Spearman's rank correlation ($\rho$), Relative $\ell2$-distance ($R_{\ell2}$), and Average Intersection over Union (AIoU) \cite{art-16}.
\subsection{Implement Details}
\noindent We adopt I3D \cite{art-13} as the backbone, which is optimized via NAdam \cite{art-18} on an NVIDIA A40 GPU. The 2D pose detector used in this paper is based on YOLOv11. Learning rates are initialized at $10^{-4}$ for the backbone and $10^{-3}$ for other modules. To implement physical constraints, we select the 12 joint points in the COCO standard, excluding the head, as our $Y$. The fusion weights for motion and condition streams are set to $\alpha=0.7$ and $\beta=0.3$. Experiments are conducted on FineDiving \cite{art-16}, FineDiving-HM \cite{art-12}, and MTL-AQA \cite{art-15} following official protocols. During inference, we employ a difficulty-aware voting strategy with $L=5$ reference samples. The average inference time is 1.39s per video on an RTX 3090.
\subsection{Comparison with the State-of-the-Arts}
\noindent We compare our proposed framework with state-of-the-art AQA methods on the FineDiving, FineDiving-HM, and MTL-AQA datasets. As shown in Tables \ref{tab:ex-finediving}, \ref{tab:ex-finediving-hm} and \ref{tab:ex-mtl}, our method consistently outperforms existing baselines across all metrics, validating the effectiveness of the proposed intrinsic motion distillation strategy.\\
\indent On the AQA task, our method establishes a new performance benchmark. Specifically, on the FineDiving dataset, we surpass the latest HP-MCoRe by 1\% in Spearman’s rank correlation ($\rho$) and reduce the Relative $\ell_2$-distance by 0.019. A remarkable finding is that our method, without using any mask annotations on FineDiving ($\rho=0.9465$), outperforms the mask-dependent state-of-the-art FineParser on FineDiving-HM ($\rho=0.9424$). This evidence strongly suggests that our pose-guided hard-attention mechanism captures the essence of action execution more effectively than expensive pixel-level foreground segmentation. Furthermore, on FineDiving-HM, where both methods utilize masks for fair comparison, we further lead by 0.79\%, proving that our explicit geometric modeling provides complementary gains beyond visual background removal. On MTL-AQA, we similarly achieve a 0.83\% gain in $\rho$ over the best competitor.\\
\indent On the Action Segmentation task, our method achieves significant improvements. By leveraging pose topology as a stable geometric prior, we improve AIoU@0.75 by 1.23\% and 4.34\% over the state-of-the-art on FineDiving and FineDiving-HM, respectively. This indicates that structural geometric features are significantly more robust than appearance-based features for detecting precise action boundaries in cluttered environments.
\subsection{Ablation Study}
\noindent We conduct comprehensive ablation studies on the FineDiving dataset to validate the effectiveness of the proposed physical constraints.
\paragraph{\textbf{Effectiveness of Spatial Hard-Attention}} Table \ref{tab:main-ablation} details the contribution of each module. We observe that Method A (baseline without decoupling and pose) yields suboptimal results. The most significant performance leap occurs in Method D, where the introduction of Action-Unit images (Hard Attention) boosts Spearman’s correlation from 0.9301 to 0.9378 and reduces Relative $\ell_2$-distance to 0.2665. This empirical evidence confirms our core hypothesis: physically filtering out background clutter via Action Units forces the model to focus on intrinsic motion, significantly enhancing representation quality. Furthermore, Method E (Pure-Pose Feature) further improves accuracy compared to Method C, proving that explicit geometric modeling captures joint-level nuances that visual features alone miss.
\paragraph{\textbf{Role of Geometric Priors in Segmentation}} Table \ref{tab:aiou-ablation} demonstrates that introducing pose topology as a geometric prior enables the Action Segmentation Module to surpass the state-of-the-art HP-MCoRe and approach human-level accuracy. Even without pose, our decoupled architecture outperforms baselines, validating the necessity of separating motion from environmental conditions. Additionally, Table \ref{tab:pure-pose} confirms that the cascaded LSTM-Transformer architecture in the Pose Extractor optimally models both local temporal dynamics and global dependencies.
\begin{table}[t]
	\centering
	\caption{Ablation study on Action Segmentation Module on FineDiving dataset.}
	\begin{tabular}{c|c|cc}
		\toprule
		Methods & Pose information & AIoU@0.5$\uparrow$ & AIoU@0.75$\uparrow$ \\
		\midrule
		G & & 0.9901 & 0.9608 \\
		\rowcolor[rgb]{.91, .91, .91} \textbf{F} & \checkmark & \textbf{0.9998} & \textbf{0.9841} \\
		\bottomrule
	\end{tabular}
	\label{tab:aiou-ablation}
\end{table}
\begin{table}[t]
	\centering
	\caption{Ablation study on Pure-Pose Feature Extractor on FineDiving dataset.}
	\begin{tabular}{c|cc|cc}
		\toprule
		Methods & LSTM & Transformer & $\rho\uparrow$ & $R_{\ell2}\downarrow(\times100)$ \\
		\midrule
		H & \checkmark & & 0.9364 & 0.2914 \\
		I & & \checkmark & 0.9358 & 0.3115 \\
		\rowcolor[rgb]{.91, .91, .91} \textbf{F} & \checkmark & \checkmark & \textbf{0.9465} & \textbf{0.2243} \\
		\bottomrule
	\end{tabular}
	\label{tab:pure-pose}
\end{table}
\paragraph{\textbf{Hyperparameters}} We investigate the balance between motion and condition features in Table \ref{tab:param-ratio}, finding that a weight ratio of $\alpha=0.7$ (motion) and $\beta=0.3$ (condition) yields the best results, reflecting the dominance of technical execution in scoring. The voting strategy is optimized at $L=5$ (Table \ref{tab:param-l}).
\begin{table}[t]
	\centering
	\caption{Impact of the number of voting samples on effectiveness under the voting sample selection strategy proposed in this paper.}
	\begin{tabular}{c|cc}
		\toprule
		$L$ & $\rho\uparrow$ & $R_{\ell2}\downarrow(\times100)$ \\
		\midrule
		1 & 0.9401 & 0.2427 \\
		\rowcolor[rgb]{.91, .91, .91} \textbf{5} & \textbf{0.9465} & \textbf{0.2243} \\
		10 & 0.9460 & 0.2285 \\
		15 & 0.9462 & 0.2251 \\
		\bottomrule
	\end{tabular}
	
	\label{tab:param-l}
\end{table}
\begin{table}[t]
	\centering
	\caption{Effects of varying the ratio between motion and condition parts on performance on the FineDiving diving dataset.}
	\begin{tabular}{c|c|cc}
		\toprule
		$\alpha$ & $\beta$ & $\rho\uparrow$ & $R_{\ell2}\downarrow(\times100)$ \\
		\midrule
		0.1 & 0.9 & 0.9421 & 0.2374 \\
		0.3 & 0.7 & 0.9429 & 0.2311 \\
		0.5 & 0.5 & 0.9457 & 0.2272 \\
		\rowcolor[rgb]{.91, .91, .91} \textbf{0.7} & \textbf{0.3} & \textbf{0.9465} & \textbf{0.2243} \\
		0.9 & 0.1 & 0.9430 & 0.2295 \\
		\bottomrule
	\end{tabular}
	\label{tab:param-ratio}
\end{table}
\paragraph{\textbf{Cross-Dataset Generalization}} To validate that our framework captures universal motion patterns rather than overfitting to dataset-specific background biases, we conduct a cross-domain evaluation by training on MTL-AQA and testing on FineDiving. As shown in Table \ref{tab:cross-experiment}, our method maintains high scoring correlation even on unseen data. This confirms that the pose-guided hard-attention mechanism effectively extracts intrinsic motion representations that remain invariant across different competition environments and camera views \cite{art-58}.
\subsection{Visualization}
\begin{figure}[!tb]
	\centering
	\includegraphics[width=1.0\linewidth]{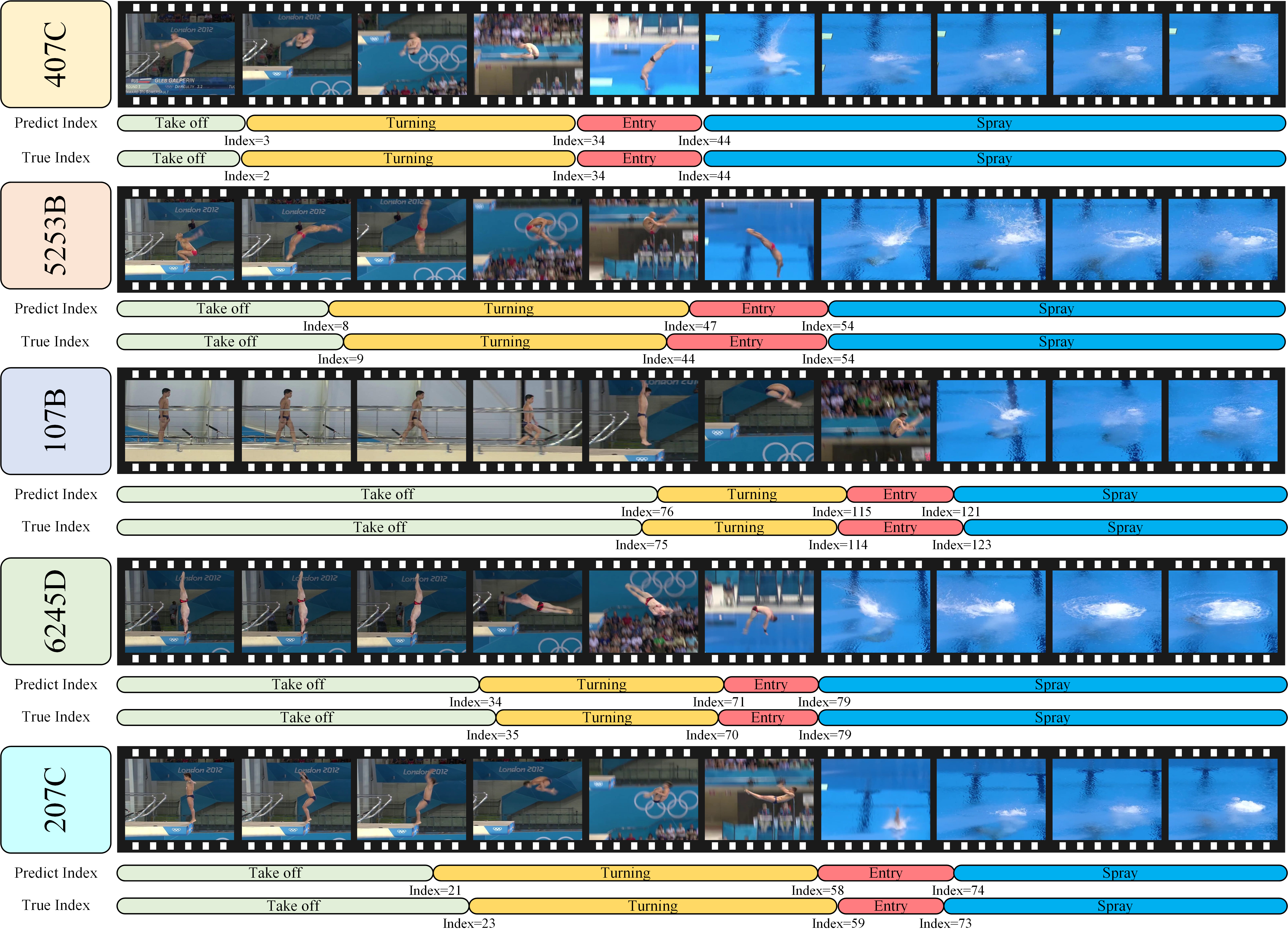}
	\caption{
		Visualization of the segmentation outputs from the Action Segmentation Module on videos representing five distinct action types.
	}
	\label{fig:visualiation}
\end{figure}

\begin{figure}[!t]
	\centering
	\includegraphics[width=1.0\linewidth]{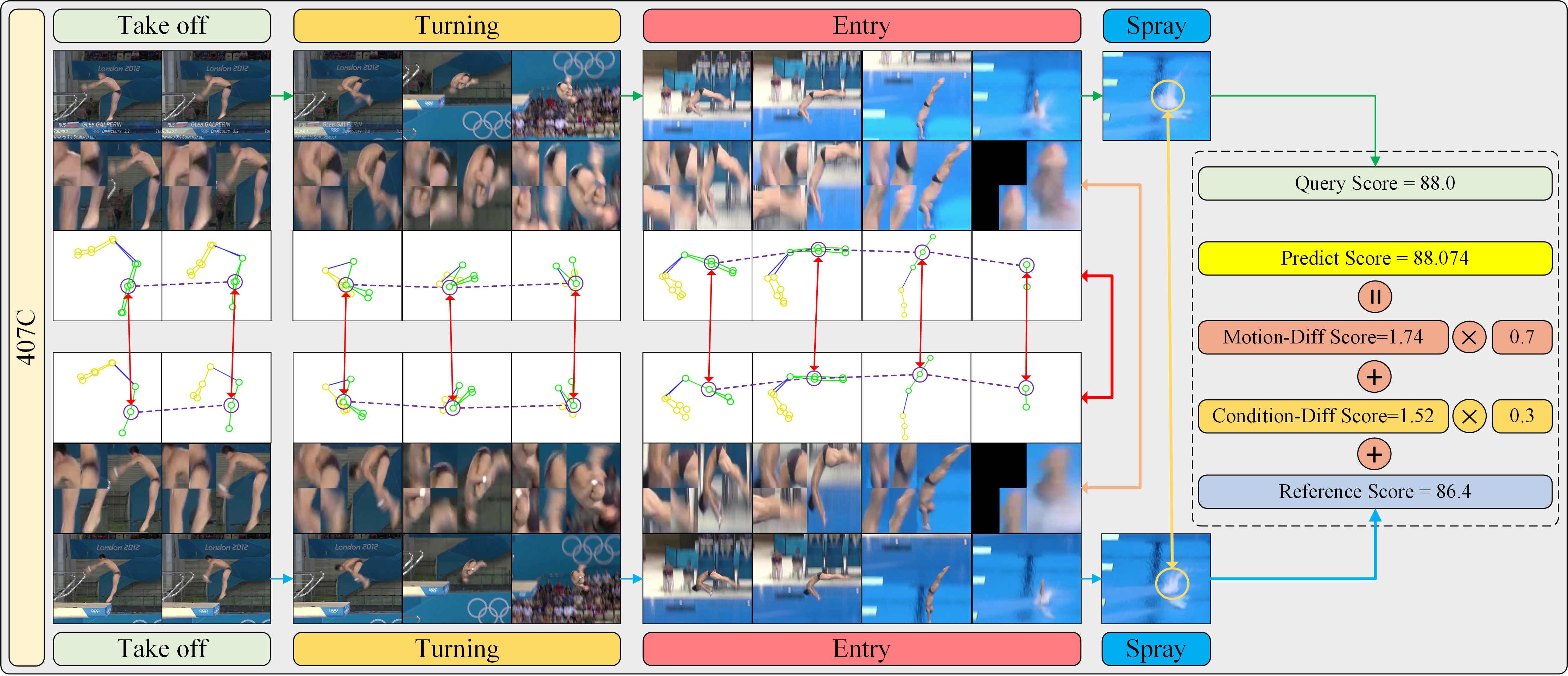}
	\caption{
		Visualization of the interpretative scoring process in our framework. Taking the ``407C'' dive as an example, the model compares a query video (top) with a reference video (bottom). It extracts pure Action Units and topological skeletons across aligned sub-phases (Take-off, Turning, Entry, and Spray) to compute the Motion-Diff Score. Simultaneously, the spray features are compared to obtain the Condition-Diff Score. The final prediction fuses these differences with learned weights ($\alpha=0.7$ for motion, $\beta=0.3$ for condition), demonstrating strong physical interpretability.
	}
	\label{fig:scoring-vis}
\end{figure}

\noindent To qualitatively validate the robustness of our framework against environmental noise, we visualize the temporal segmentation results for five distinct diving actions (407C, 5253B, 107B, 6245D and 207C) in Figure \ref{fig:visualiation}. As observed, the model precisely identifies the boundaries of four critical stages: take-off, turning, entry, and spray. Even in frames with significant occlusion (e.g., water splash) or rapid transitions, the predicted boundaries align almost perfectly with expert annotations (Ground Truth). This visualization confirms that our pose-guided hard-attention mechanism effectively filters out background clutter, enabling the model to capture the intrinsic temporal structure of complex actions with human-level accuracy.
\indent Furthermore, to demonstrate the interpretability of our scoring mechanism, we visualize the detailed score inference process for the ``407C'' action in Figure \ref{fig:scoring-vis}. As illustrated, our Align Module successfully divides the sequence into four distinct sub-phases (Take-off, Turning, Entry, and Spray). By physically extracting Action Units and topological skeletons, the model performs fine-grained comparisons between the query and reference videos strictly within the body regions, effectively isolating the motion execution (Motion-Diff) from environmental outcomes like water splash (Condition-Diff). The final score is transparently aggregated using our Weight-Adjust Scoring Module, proving that our framework assesses action quality with high physical interpretability rather than relying on black-box visual shortcuts.
\section{Conclusion}
\noindent In this paper, we address the critical challenge of low signal-to-noise ratio in AQA by proposing a Pose-Guided Intrinsic Motion Distillation Framework. Unlike previous approaches that struggle with background clutter, we introduce Action Units as a spatial hard-attention filter, effectively enforcing physical constraints to isolate intrinsic motion from visual noise. Furthermore, our orthogonal decoupling strategy ensures independent and fair evaluation of technical execution and environmental conditions. Extensive experiments on large-scale benchmarks demonstrate that our method sets a new state-of-the-art in both action segmentation and scoring. These results underscore the importance of physically grounded feature extraction—rather than purely data-driven learning—for achieving robust and interpretable action quality assessment. 
\section{Limitations and Future Work}
\noindent To efficiently validate the efficacy of incorporating spatial geometric priors, our current framework relies on an off-the-shelf 2D pose estimator. Empirically, even when the estimator yields sub-optimal joint coordinates under extreme conditions (e.g., severe motion blur, atypical body distortions, or low-resolution inputs), our decoupled architecture maintains strong empirical robustness, consistently outperforming existing baselines. However, given the absence of ground-truth pose annotations in existing public AQA datasets, it remains challenging to quantitatively isolate the causal direct impact of pose estimation accuracy on the final scoring performance. Recent advances in differentiable pose estimation and weakly-supervised learning offer promising avenues to alleviate this annotation bottleneck. Future work aims to explore the end-to-end joint optimization of pose extraction and action evaluation to mitigate this potential error cascading.\\
\indent Moreover, relying solely on 2D pose estimation inherently limits the model's ability to handle severe self-occlusions and complex out-of-plane rotations typical in high-diving actions. In our future work, we plan to extend the Action-Unit Parser by integrating 3D human pose reconstruction techniques. Extracting robust 3D kinematic features not only resolves depth ambiguities but also provides a more comprehensive geometric representation of the athlete's spatial dynamics. We believe that shifting from 2D planar constraints to 3D spatiotemporal modeling serves as a critical step toward achieving human-expert-level action quality assessment in unconstrained environments.
%% Add \usepackage{lineno} before \begin{document} and uncomment 
%% following line to enable line numbers
%% \linenumbers

%% main text
%%

%% Use \section commands to start a section

%% For citations use: 
%%       \citet{<label>} ==> Lamport [21]
%%       \citep{<label>} ==> [21]
%%

%% 使用 BibTeX
\bibliographystyle{elsarticle-num-names} 
\bibliography{references}  % 注意：这里不要加 .bib 扩展名

%% If you have bib database file and want bibtex to generate the
%% bibitems, please use
%%
%%  \bibliographystyle{elsarticle-num-names} 
%%  \bibliography{<your bibdatabase>}

%% else use the following coding to input the bibitems directly in the
%% TeX file.

%% Refer following link for more details about bibliography and citations.
%% https://en.wikibooks.org/wiki/LaTeX/Bibliography_Management

\end{document}